\documentclass[conference]{IEEEtran}

\usepackage[T1]{fontenc}
\usepackage{balance}
\usepackage{comment}
\usepackage{dirtree}
\usepackage{float}
\usepackage{fontawesome}
\usepackage{graphicx}
\usepackage{hhline}
\usepackage{listings}
\usepackage{multirow}
\usepackage{xspace}
\usepackage{subcaption}
\usepackage{fancyvrb}
\usepackage{adjustbox}
\usepackage{mdframed}
\usepackage{url} 
\usepackage{tabularx}

\usepackage[acronym]{glossaries}
\newacronym{ai}{AI}{artificial intelligence}
\newacronym{ar}{AR}{augmented reality}
\newacronym{gui}{GUI}{Graphical User Interface}
\newacronym{hls}{HLS}{HTTP Live Streaming}
\newacronym{iou}{IoU}{Intersection-over-Union}
\newacronym{mot}{MOT}{Multi-Object-Tracking}
\newacronym{ocr}{OCR}{Optical Character Recognition}
\newacronym{reid}{Re-ID}{Re-identification}

\newcommand{\frameworkname}[0]{PlayerTV\xspace}

\begin{document}
\bstctlcite{IEEEexample:BSTcontrol}

\title{\frameworkname: Advanced Player Tracking and Identification for Automatic Soccer Highlight Clips}

\author{\IEEEauthorblockN{
Håkon Maric Solberg\IEEEauthorrefmark{1},
Mehdi Houshmand Sarkhoosh\IEEEauthorrefmark{2}\IEEEauthorrefmark{4},
Sushant Gautam\IEEEauthorrefmark{2}\IEEEauthorrefmark{3}, \\
Saeed Shafiee Sabet\IEEEauthorrefmark{4}, 
Pål Halvorsen\IEEEauthorrefmark{1}\IEEEauthorrefmark{2}\IEEEauthorrefmark{3}\IEEEauthorrefmark{4}, 
Cise Midoglu\IEEEauthorrefmark{3}\IEEEauthorrefmark{4}}
\IEEEauthorblockA{\IEEEauthorrefmark{1}University of Oslo, Norway}
\IEEEauthorblockA{\IEEEauthorrefmark{2}Oslo Metropolitan University, Norway}
\IEEEauthorblockA{\IEEEauthorrefmark{3}SimulaMet, Norway}
\IEEEauthorblockA{\IEEEauthorrefmark{4}Forzasys AS, Norway}
}

\maketitle

\begin{abstract}
In the rapidly evolving field of sports analytics, the automation of targeted video processing is a pivotal advancement. We propose \frameworkname, an innovative framework which harnesses state-of-the-art AI technologies for automatic player tracking and identification in soccer videos. By integrating object detection and tracking, \gls{ocr}, and color analysis, \frameworkname facilitates the generation of player-specific highlight clips from extensive game footage, significantly reducing the manual labor traditionally associated with such tasks. Preliminary results from the evaluation of our core pipeline, tested on a dataset from the Norwegian Eliteserien league, indicate that \frameworkname can accurately and efficiently identify teams and players, and our interactive \gls{gui} serves as a user-friendly application wrapping this functionality for streamlined use.
\end{abstract} 


\IEEEpeerreviewmaketitle

\section{Introduction}\label{section:introduction}

Sport video content is consumed in various ways, both for entertainment and for analysis~\cite{sarkhoosh2023soccer}. In this respect, a lot of manual tedious work is required to prepare and compile compelling and usable summaries of the potentially long video assets. Modern technologies enable automatic computer vision and \gls{ai} based solutions to make more efficient production pipelines~\cite{Midoglu2024Feb}, e.g., using automatic event detection~\cite{nergardrongvedAutomatedEventDetection2021}, clipping~\cite{Valand++21}, video aspect ratio retargeting (cropping)~\cite{SmartCrop_Majidi}, and using the available multimodal data to create text summaries~\cite{Gautam2022Oct}. 

Using soccer as an example, we aim to push the automation one step further and allow the creation of video clips which focus on individual athletes. In this context, the development and use of advanced player tracking technologies is emerging as a cornerstone to enhance tournament strategies, player performance, game analytics and fan engagement. With our proposed framework \frameworkname, we deploy state-of-the-art AI models to track the players in video frames, perform player identification through team detection and kit number identification (relying on color analysis and \gls{ocr}, resolving occlusion and collision situations in the process), and clip parts of the input video, to extract frames containing the target player, i.e., generating a sophisticated video processing pipeline to enable player specific soccer highlight clips. We hope that our proof-of-concept demonstration of the technological advantages of such a framework can open up new avenues for applied research, especially in the areas of computer vision for player performance analysis, automatic multimedia content for fan experiences and highlights. Our contributions are as follows:

\begin{itemize}

    \item We present a framework called \frameworkname for automatically creating per player highlight clips from input videos using object detection and tracking, \gls{ocr}, unsupervised clustering, and database lookup. The core pipeline comprises 5 modules (player tracking, RGB, \gls{ocr}, team mapping, player mapping) and is highly configurable.
    
    \item An interactive \gls{gui} serves as a wrapper around the core pipeline, making \frameworkname a user-friendly application where users can specify an input video and configuration parameters, and retrieve a clipped video including only the scenes of a selected player. 
    
    \item We provide an open source implementation for the entire framework, along with additional artifacts. 
    
    \item We present initial experiment results serving as a preliminary evaluation of the performance of the pipeline under different configurations.
    
\end{itemize}

\section{Background and Related Work}\label{section:background}


\subsection{Sports Analytics}

The domain of sports analytics has witnessed substantial growth, focusing primarily on enhancing team performance, refining strategic approaches, and improving player evaluation~\cite{garganta2009trends, Rein2016Aug}. Professional sports teams invest significantly in analytical tools to gain competitive advantages~\cite{Pizzo2022Feb, Torres-Ronda2022Jan}. Sports analytics platforms~\cite{Wyscout, SportsBase} are extensively utilized in professional soccer for detailed player and game analysis. Currently, much of the data annotation involved in sports analytics remains manual, a labour-intensive process that limits the speed and scalability of data utilization~\cite{Liu2023Nov, Midoglu2022Feb}. There is a compelling case for automation in this field, which could not only streamline existing processes but also enable advanced, real-time analytical capabilities, thus transforming the landscape of sports analytics~\cite{Patel2020Mar}. Other interesting research includes spotting key game events such as goals or fouls~\cite{Morra2020Jun, Rongved}, summarizing games automatically~\cite{Gautam2022Oct}, multimodal game understanding~\cite{gautam2022assisting,Gautam2023Oct}, creating exciting highlights~\cite{Midoglu2024Feb}, and generating captions for game footage~\cite{Mkhallati, Qi2023Oct}.


\subsection{Object Detection and Tracking}

In the realm of sports analytics, object detection and tracking algorithms are indispensable for the automatic localization and movement analysis of players and balls across video frames~\cite{Naik2022Apr}. YOLO~\cite{yolo-main} excels in high-speed detection through its unique approach to processing images in a single evaluation, making it especially suitable for dynamic sporting events. SSD~\cite{SSD-paper}, on the other hand, achieves a commendable balance between speed and accuracy by employing a deep neural network to detect multiple objects simultaneously. For object tracking, SORT~\cite{sort-tracking} and DeepSORT~\cite{deepsort} offer robust solutions, enhancing tracking consistency by integrating deep learning to manage the dynamic nature of sports environments. The Deep-EIoU~\cite{deepEIOU} algorithm improves tracking precision through iterative alignment using an advanced \gls{iou} operation and deep features, essential for handling fast-moving and overlapping objects. These technological advancements are pivotal in enhancing the automation of data annotation and enabling sophisticated real-time analysis~\cite{Pavitt2021Mar}, thereby aiding in the formulation of immediate and strategic decisions in sports analytics.


\subsection{Optical Character Recognition (OCR)}

\gls{ocr} converts different types of documents into editable and searchable data and has significantly advanced due to improvements in machine learning and image processing~\cite{Memon2020Jul}. In sports analytics, \gls{ocr} is utilized to analyze game footage by extracting jersey (kit) numbers and statistics from scoreboards~\cite{Alhejaily2023Mar, Sharma2024Apr,Cioppa_2022_CVPR, Shih2017Jan}. This technology automates real-time decision-making and historical data analysis, proving vital for coaching strategies and performance assessment~\cite{Seweryn2023Sep}. Future developments in \gls{ocr} may focus on increasing accuracy in dynamic environments~\cite{Hamdi2023Mar}, offering profound impacts on sports analytics and media broadcasting.


\subsection{Player Tracking in Sports}

The evolution of player tracking technologies in sports, particularly in soccer, illustrates a significant technological advancement in the realm of sports analytics~\cite{deepEIOU}. These technologies have transitioned from basic manual methods to highly sophisticated systems, encompassing both optical tracking and sensor-based systems~\cite{Seckin2023Sep}. Optical tracking systems utilize high-definition cameras positioned around the playing field to capture comprehensive visual data~\cite{Liu2023Nov}, which is then processed to track player and ball movements across the pitch. Sensor-based systems, on the other hand, involve the use of wearable technologies embedded in player gear, providing real-time data on player positions, movements, and even biometrics~\cite{DeFazio2023Feb}.

\textbf{Importance of tracking information:} Accurate game data, such as player positions and ball location, is indispensable in modern sports science~\cite{Rein2016Aug}. This data underpins not only tactical and strategic decisions made by coaching staff but also enhances sports science research focused on performance enhancement and injury prevention~~\cite{Rein2016Aug, Lord2020Oct}. Furthermore, tracking information enriches fan engagement by offering detailed game analytics, contributing to more immersive viewing experiences and interactive fan involvement~\cite{Wu2022May}. Through the use of comprehensive tracking data, teams can optimize their strategies, improve player conditioning programs, and enhance the overall spectator experience by providing insights that were previously inaccessible.

\textbf{Challenges in player tracking:} Accurately tracking players in sports settings presents several distinct challenges~\cite{Torres-Ronda2022Jan, Naik2022Apr}. First, occlusion issues arise frequently, where players or the ball are obscured by other players, leading to temporary loss of tracking data~\cite{Liu2023Nov}. Secondly, varying lighting conditions, such as shadows or stadium lights, can significantly alter the visibility and appearance of players~\cite{Rahimian2022Mar, Naik2022Apr}, complicating the tracking process. Additionally, the fast-paced and unpredictable movements typical of sports scenarios challenge the tracking algorithms’ ability to maintain consistent tracking. These movements often result in motion blur, which can degrade the quality of the visual data used for tracking. State-of-the-art tracking algorithms, such as those employing Kalman filters, are designed for objects moving in smooth, predictable trajectories and thus struggle with the erratic movement patterns found in sports~\cite{Liu2023Nov, deepsort, sort-tracking, aharon2022botsort, cao2023observationcentric}. Research in player \gls{reid} techniques focuses on overcoming the challenges posed by changes in appearance or occlusions in sports environments, aiming to accurately track and identify players across different situations~\cite{cioppa2023soccernet}. These studies delve into innovative methods to enhance recognition accuracy, crucial for various applications including player performance analysis and sports event security.

\textbf{Data integration and visualization}: Previous efforts have centered on seamlessly integrating tracking data with game metadata, enabling comprehensive analysis and enhancing broadcast experiences~\cite{cioppa2023soccernet, Cioppa_2022_CVPR}. By visualizing this integrated information, stakeholders can gain deeper insights into player movements, team strategies, and overall game dynamics, fostering a richer understanding of sporting events~\cite{SmartCrop_Majidi, Gautam2023Oct}. 

Improvements in all the underlying technologies point towards the transformative potential of advanced player tracking and AI-driven analysis in the realm of sports, especially soccer, by refining tactics and enriching fan experiences. However, challenges like occlusion, varying lighting, and unpredictable player movements still hinder the accuracy of these technologies. Our aim is to overcome these limitations by developing innovative methods that enhance the precision and functionality of player tracking, paving the way for advanced sports analytics solutions.

\begin{figure*}[!h]
    \centering
    \includegraphics[width=0.95\textwidth]{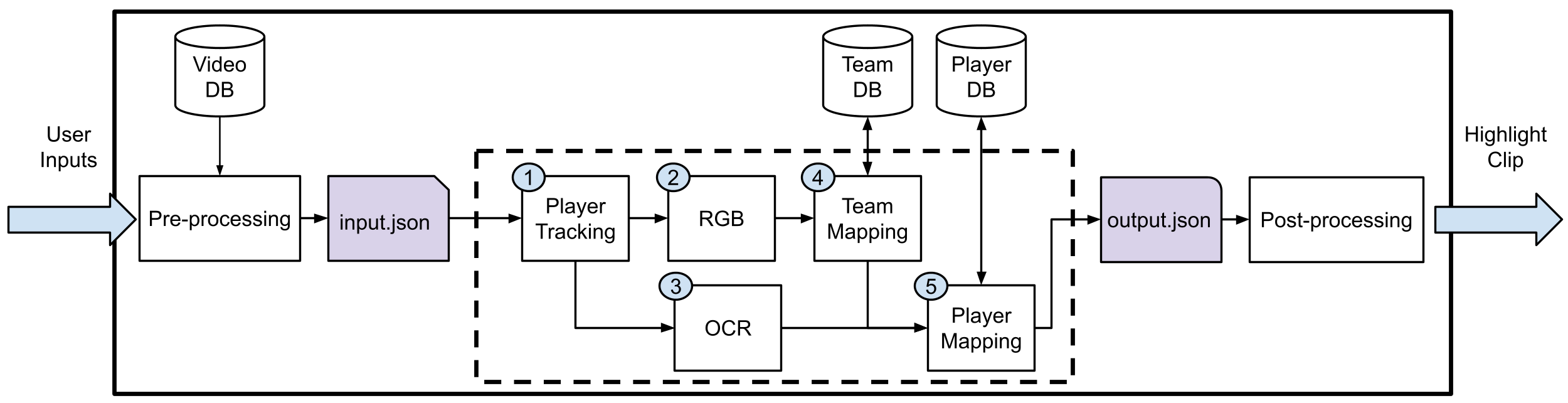}
    \caption{\frameworkname framework overview.}
    \label{fig:playertv-overview}
\end{figure*}


\subsection{Motivation and Potential Use Cases for \frameworkname}

There are various use cases for the \frameworkname framework: 
\begin{itemize}

    \item \textbf{Tactical analysis}: coaches and analysts can use \frameworkname for in-depth tactical analysis, identifying patterns in player movements, and optimizing team strategies.
    
    \item \textbf{Performance analysis and player development}: tracking data can help in evaluating player performances, designing personalized training programs, and monitoring progress over time.

    \item \textbf{Broadcast enhancements}: tracking data can be incorporated to provide viewers with enriched content, such as live player statistics, heat maps, and tactical overlays.

    \item \textbf{Fan engagement and interactive experiences}: tracking data can be used to create immersive fan experiences, such as \gls{ar} applications that overlay player stats and movements in real-time during a match.

    \item \textbf{Betting and fantasy sports}: tracking data can provide insights for betting odds and fantasy sports, offering detailed analytics to users for making informed decisions.

    \item \textbf{Injury prevention and management}: analyzing player movements and workload can aid in identifying potential injury risks and managing player fitness levels.

\end{itemize}


\section{Proposed Framework}\label{section:framework}

The \frameworkname framework consists of a core pipeline and an interactive \gls{gui} that serves as a wrapper around the core pipeline. Figure~\ref{fig:playertv-overview} presents an overview of the framework, where the dashed line indicates the core pipeline and the solid line indicates the wrapper.


\subsection{Core Pipeline}

The \frameworkname core pipeline comprises 5 modules, takes a JSON configuration file as input, and outputs a JSON file.


\subsubsection{Player Tracking Module}

\begin{figure}[htbp]
    \centering
    \includegraphics[width=0.9\columnwidth]{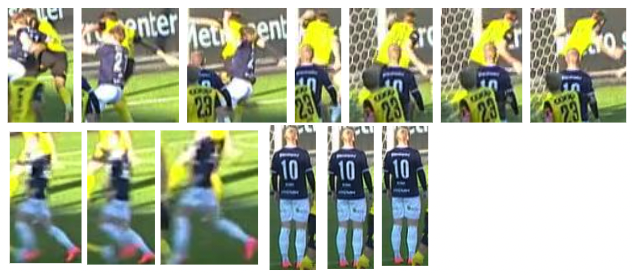}
    \caption{Sample tracklets. Both noise from other players as well as occlusion makes it difficult to correctly detect kit numbers.}
    \label{fig:sample-tracklets}
\end{figure}

For tracking different players across between frames, we have taken the Deep-EIoU tracker~\cite{deepEIOU} as a starting point. This tracker has been shown to outperform DeepSORT~\cite{deepsort}, BoT-SORT~\cite{aharon2022botsort} and OC-SORT~\cite{cao2023observationcentric} for \gls{mot} in sports. The tracking module supports using a YOLOX~\cite{yolox} model that has been trained on the SoccerNet tracking dataset~\cite{cioppa2023soccernet}, as well as SportMOT~\cite{cui2023sportsmot}. Figure~\ref{fig:sample-tracklets} presents sample "tracklet"s (crops of detected players), which are still challenging to use for direct player identification due to noise and occlusions. We have modified the tracker to also extract \gls{iou} and BRISQUE-score~\cite{brisque} from the tracklets. In addition to this, we also keep an average color value from a region of interest within each crop. These are later used for mapping players to teams. 

\subsubsection{RGB Module}

\begin{figure}[htbp]
    \centering
    \includegraphics[width=0.75\columnwidth]{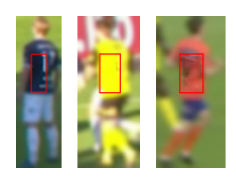}
    \caption{Sample regions of interest within tracklets.}
    \label{fig:rgb-crop}
\end{figure}

The RGB information retrieved from the region of interest within the tracklets are kept for later clustering into the 2 different teams. As there is a lot of grass (green) surrounding each player, this would blend out the colors, as we are trying to capture values closest to the players kit color. By using an offset where we only look at the top half, while squeezing in the sides and top of the detection, we retrieve a more distinct color, creating more distance between samples of different teams. Figure~\ref{fig:rgb-crop} presents examples of these regions of interest. The offset proved to be most effective for clustering, is a clipping removing $\frac{1}{3}$ of the crop on both sides, $\frac{1}{4}$ of the top and $\frac{1}{2}$ of the bottom.

All RGB values for detections with no intersections to other detections are used as a sample base. After normalizing, then clustering, we use the distance from cluster centers to map tracked player to the correct team, unless their RGB were used for clustering itself. Team home color is retrieved from the PLAYER DB, and the cluster center closest to it is mapped accordingly.

\begin{figure}[htbp]
    \centering
    \includegraphics[width=0.8\columnwidth]{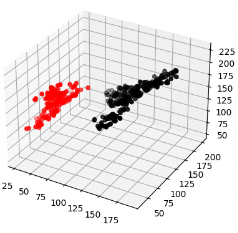}
    \caption{Sample clustering. Illustrated in RGB space.}
    \label{fig:sample-clustering}
\end{figure}

When extracting the team colors we provide 2 options for the color space: CIELAB\footnote{\url{https://en.wikipedia.org/wiki/CIELAB_color_space}} and weighted RGB. CIELAB is a color space consisting of L* for lightness, a* from green to red, and b* from blue to yellow. This color space is designed to better mimic human perception. A rough estimation to this is using the weighted RGB, where the rgb channels are weighted 0.3, 0.59, and 0.11, respectively. In this stage, an \gls{iou} threshold can also be introduced, where only the tracklets with low \gls{iou} (for which the sum of all overlaps with other tracklets are below the specified threshold) are considered, instead of all tracklets. Then, we run k-means clustering with k=2 (Figure~\ref{fig:sample-clustering}).

\subsubsection{OCR Module}

An additional step for identifying the player is to extract the kit number. This module supports using two alternative methods on detected crops: PaddleOCR~\cite{paddle} or EasyOCR~\cite{EasyOCR}. To increase the probability of getting good results from the OCR, we run through the best frames for each tracked player. The scoring function for a detection \(d\) is: 

\begin{equation}
   score_d = (d.BRISQUE * d.IoU) + d.BRISQUE 
\end{equation}

A lower BRISQUE score is associated with a better image quality, we therefore try minimize this value. When there is no overlap with another detection in the frame, only BRISQUE accounts for the score. The use of \gls{ocr} could potentially confuse two detections that overlap, giving another detection's kit number as output. We therefore value a small overlapping area greatly.

When using this score function for deciding which crops to run OCR detecion on, crops close to each other tend to be high-scoring, while very similar. To increase the probability of correctly running character recognition for crops with kit number present, we divide the player's frames into different time zones, with each time zone containing the top scoring crops in descending order. This to spread out the player kit number detection in time.

By splitting a player's frames into 5 different time segments, each containing a maximum 10 proposed frames. This is to make sure we use samples spread across different time periods. When a detection is above a certain threshold, we skip the rest of the samples within that same time window. After this process we are set with a set of proposed kit numbers associated with each track ID.

In addition to splitting each player tracklet into time zones, we also provide different paradigms for running either of the \gls{ocr} model on crops:

\begin{itemize}
    \item \textbf{All tracklets, keep top 20:} This is an offline \gls{ocr} extraction method where we keep a stack of the top \textit{n} scoring crops, running \gls{ocr} after the tracker is finished.
    \item \textbf{OCR output confidence threshold:} Running \gls{ocr} on all crops with a score above a threshold.
\end{itemize}

Also a \textit{"Stop at detection"} parameter decides whether or not we want to run further kit number detection for tracklets where a detection has been found. When split into time zones, this resets for each time zone.


\subsubsection{Team Mapping Module}

In this module, home and away team information (where applicable\footnote{If the pipeline is run with an input video from our partner company Forzasys (\url{https://forzasys.com/}), the pipeline can use the corresponding game id in the M3U8 playlist URL for fetching home and away team information from the Forzasys API.}) 
are used for a kit color lookup. For this, we use a TEAM DB where we have mapped each Eliteserien team to two RGB values, corresponding to home and away kit colors, as follows: 

\begin{small}
\textless team\_name\textgreater  \textless home\_kit\textgreater 
\textless R,G,B\textgreater  
\textless away\_kit\textgreater 
\textless R,G,B\textgreater
\end{small}

The cluster closest to the home team color is then mapped accordingly, giving a tracklet its associated team name, depending on its cluster id.


\subsubsection{Player Mapping Module}

Based on the \gls{ocr} and team mapping results, we identify, per tracklet, a player entity. For this, we use a PLAYER DB where we have mapped each player in each Eliteserien team to their corresponding position and kit number within the team. 

\begin{small}
\textless team\_name\textgreater  \textless player\_name\textgreater \textless position\textgreater \textless kit\_number\textgreater
\end{small}

As there is only one kit number per player per team, collisions are handled according to team vs kit number confidence. If a kit number has higher confidence than the team affiliation, we override the team mapping if the kit number detected is unused for the other team. Vice versa, we decide kit number using a less confident kit detection if present, and the colliding tracklet is more confident. This module provides the final online team/kit affiliation per tracklet (\textit{output.json} in Figure~\ref{fig:playertv-overview}).

\begin{small}
\textless frame\_number\textgreater \textless tracker\_id\textgreater \textless x\textgreater \textless y\textgreater \textless w\textgreater \textless h\textgreater \\
\indent \textless conf\textgreater [\textless r\textgreater \textless g\textgreater \textless b\textgreater] \textless brisque\_score\textgreater \textless IoU\textgreater \textless team\_id\textgreater \\
\indent \textless kit\_number\textgreater
\end{small}

\begin{figure}[h!]
    
    \small
    \centering
    
    \begin{minipage}{1\columnwidth}
        \dirtree{%
            .1 \faFileO\space output.json .
            .2 "metadata" .
            .3 "id": <id> .
            .3 "game\_id": <game\_id> .
            .3 "clip\_id": <clip\_id> .
            .3 "width": <width> .
            .3 "height": <height> .
            .3 "fps": <fps> .
            .3 "video\_url": <video\_url> .
            .3 "api\_info" .
            .4 "home\_team": <home\_team> .
            .4 "home\_color": <home\_color> .
            .4 "away\_team": <away\_team> .
            .4 "away\_color": <away\_color> .
            .4 "home\_team\_idx": <home\_team\_idx> .
            .4 "away\_team\_idx": <away\_team\_idx> .
            .4 "home\_team\_player" .
            .5 {\faUser\space} .
            .6 "name": <player\_name> .
            .6 "shirt\_number": <shirt\_number> .
            .5 {...} .
            .4 "away\_team\_player" .
            .5 {\faUser\space} .
            .5 {...} .
            .2 "frame" .
            .3 <frame\_number> .
            .4 <track\_id> .
            .5 "frame": <frame\_number> .
            .5 "track\_id": <track\_id> .
            .5 "xywh": .
            .6 <x> .
            .6 <y> .
            .6 <width> .
            .6 <height> .
            .5 "conf": <conf> .
            .5 "avg\_rgb": <avg\_rgb> .
            .5 "brisque\_score": <brisque\_score> .
            .5 "iou": <iou> .
            .5 "team\_id": <team\_id> .
            .5 "kit\_number": <kit\_number> .
            .4 {...} .
            .3 {...} .
        }
    \end{minipage}
    
    \caption{Structure of \texttt{output.json}.}
    \label{fig:output-json}
    
\end{figure}

Figure~\ref{fig:output-json} presents the structure of the \texttt{output.json} file.


\subsection{Interactive GUI}

The \frameworkname \gls{gui} wraps around the core pipeline. Built upon the Flask web framework, this \gls{gui} provides an interactive platform for visual applications of the core pipeline. Central to its functionality, Flask plays a pivotal role in seamlessly integrating front-end user interactions with the back-end processing. This integration is key to delivering a smooth and responsive user experience. The \gls{gui}'s front-end design, created using HTML and CSS, offers an easy-to-navigate interface, thereby significantly enhancing user engagement and operational efficiency. Overall, the \gls{gui} not only enables users to run the core pipeline and clip soccer videos based on selected players but also displays an overlay indicating the location within frames where the target player has been identified.


\subsubsection{Video Input}

\begin{figure}[htbp]
    \centering
    \frame{\includegraphics[width=\columnwidth]{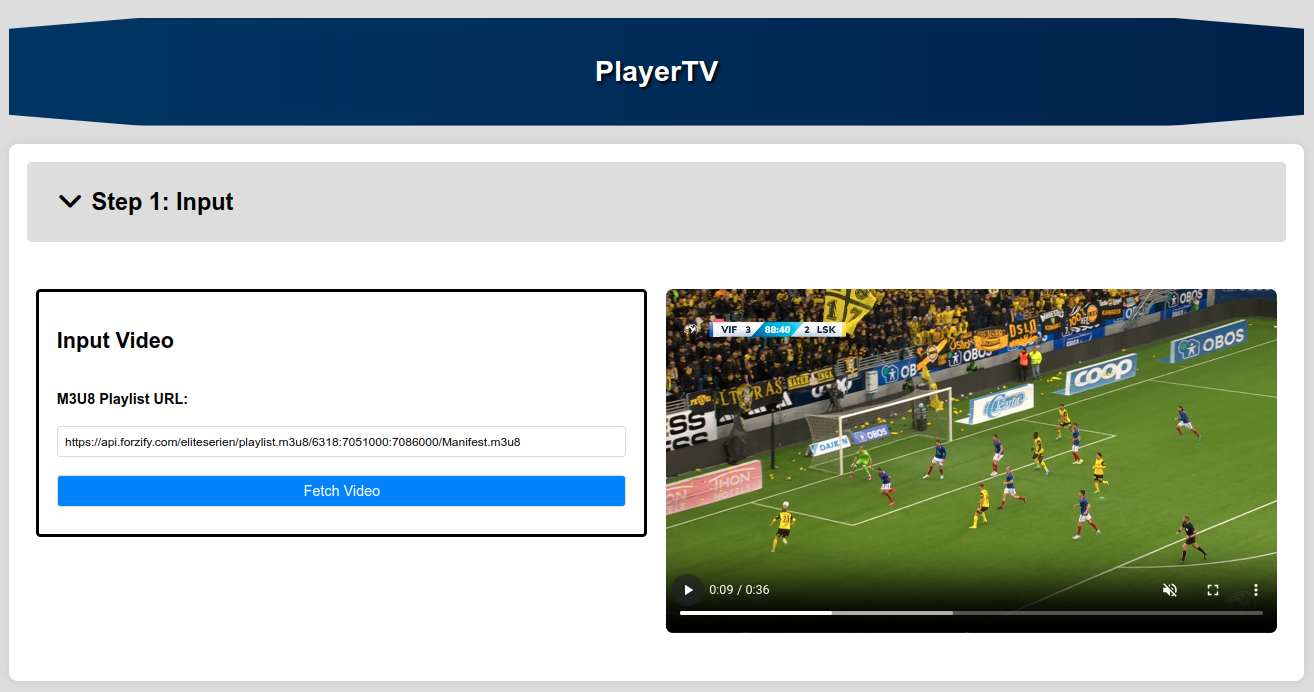}}
    \caption{\frameworkname \gls{gui} step 1.}
    \label{fig:demo-1}
\end{figure}

In the first step, \frameworkname is configured to process \gls{hls} playlists, specifically from live servers. This functionality is distinct in its capacity to handle the M3U8 format. While the overall architecture of the system, encompassing both backend and frontend, has the capability to manage local MP4 video paths, this module focuses exclusively on M3U8. The choice of this format is integral to the application's functioning, as it facilitates internal API calls. These calls play a crucial role in extracting detailed information about the soccer teams in the live stream, such as identifying home and away teams in a given match. This specific information is accessible through public APIs provided by Forzasys, which our system leverages to enhance the user experience and provide relevant context within the application (see Figure~\ref{fig:demo-1}).

\subsubsection{Configuration}

\begin{figure}[htbp]
    \centering
    \frame{\includegraphics[width=\columnwidth]{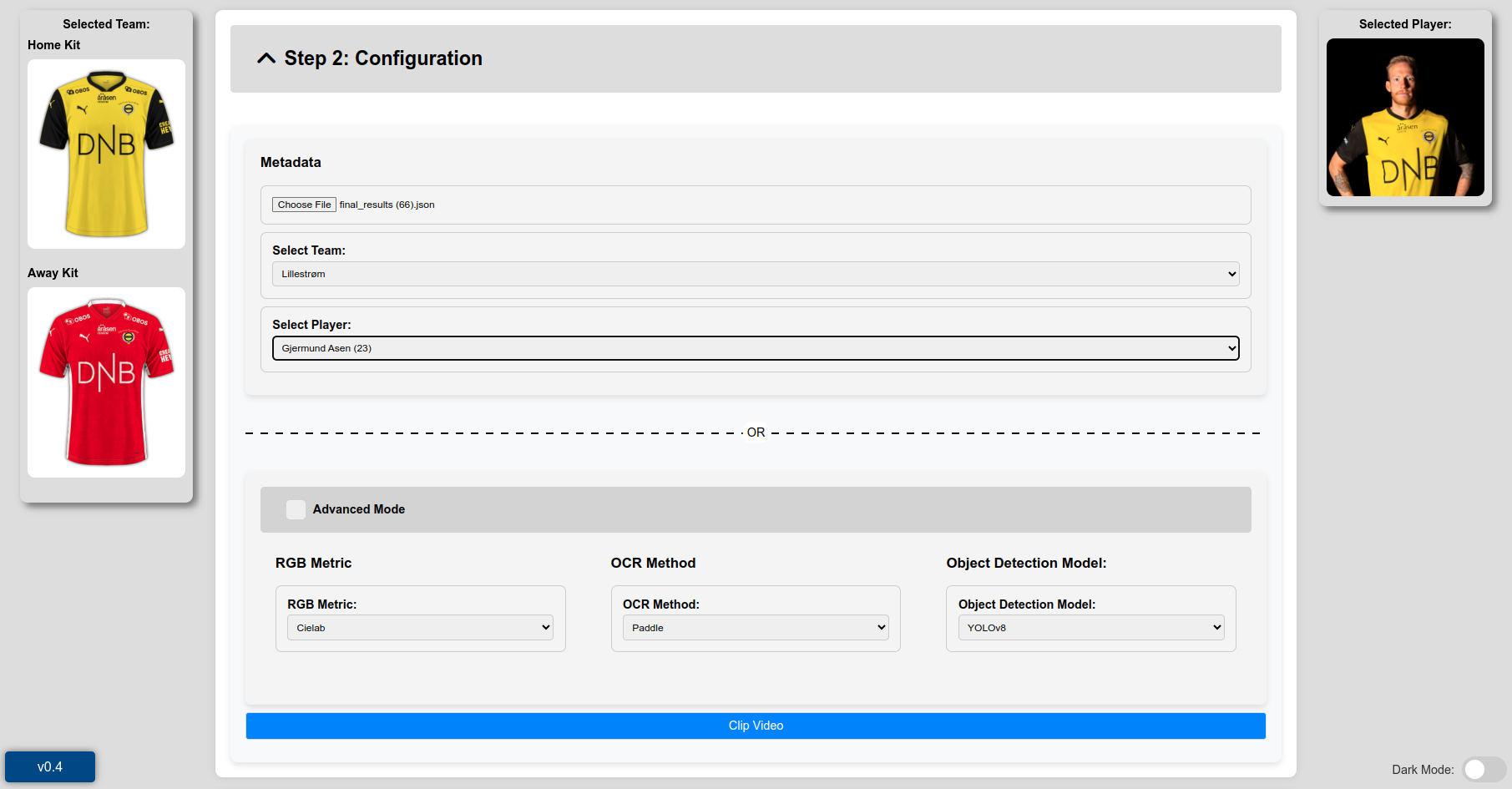}}
    \caption{\frameworkname \gls{gui} step 2.}
    \label{fig:demo-2}
\end{figure}

\begin{figure}[htbp]
    \centering
    \frame{\includegraphics[width=\columnwidth]{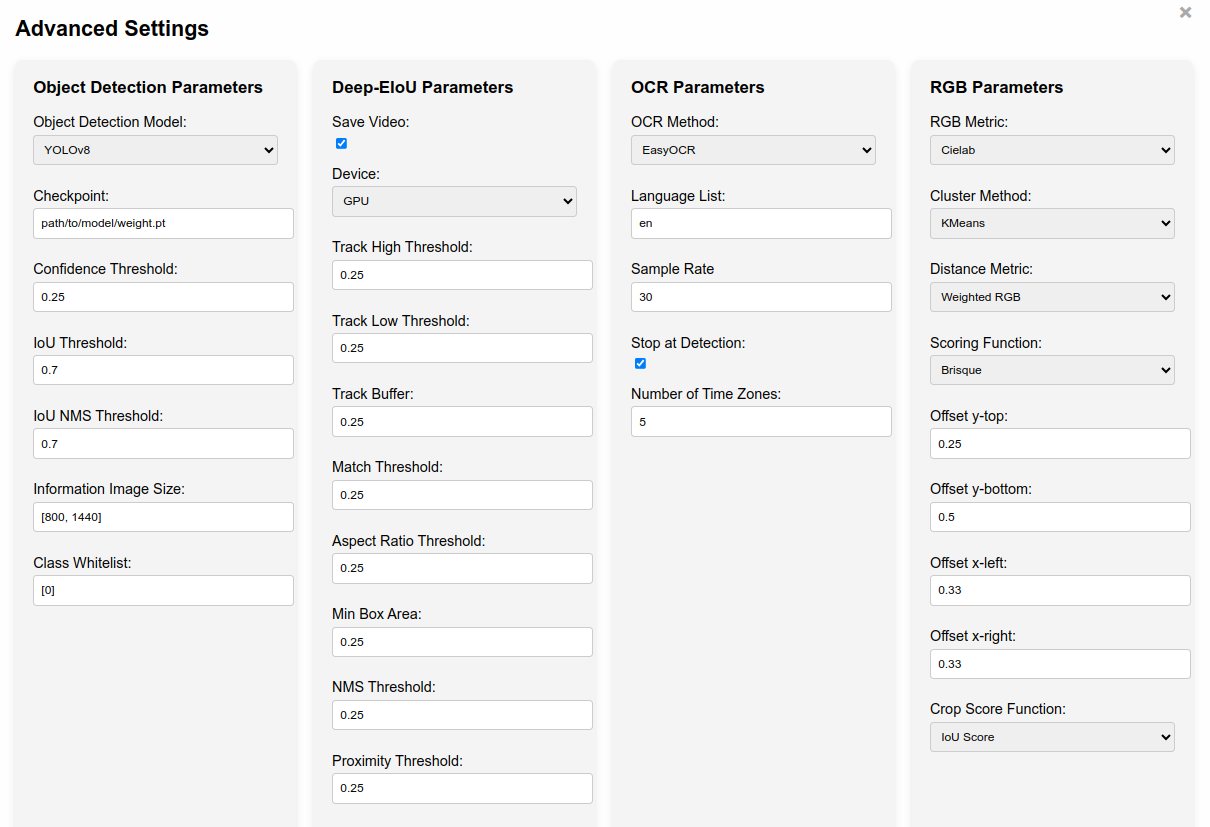}}
    \caption{\frameworkname \gls{gui} step 2 (advanced settings).}
    \label{fig:demo-2-advanced}
\end{figure}

The second step is dedicated to the configuration of the core pipeline. This section works closely with the backend to provide the user with options to select different methods and metrics for various modules. Minimal (Figures~\ref{fig:demo-2}) or advanced (\ref{fig:demo-2-advanced}) configuration options can be used, depending on whether the user wants to retain or update system defaults. Table~\ref{tab:config-parameters} presents a list of selected core pipeline configuration parameters. The parameters in the \gls{gui} serve as a user-friendly interface to set these core parameters. 

\begin{itemize}

    \item \textbf{RGB Metric:} The user has the option to choose between two metrics for clustering and identifying teams: Cielab and a weighted RGB metric. Each method focuses on clustering based on the average RGB color of the bounding boxes around detected players, helping to distinguish the home team from the away team.

    \item \textbf{OCR Method:} Users can choose between PaddleOCR and EasyOCR for the identification of players' kit numbers. These options facilitate the recognition of specific player details.

    \item \textbf{Detector:} The application offers a choice between YOLOv8 and YOLOvX models for object detection. Both models have been fine-tuned separately on a soccer dataset, encompassing various classes, with a focus on player detection.

    \textbf{Metadata:} In this section, users can select from teams and players that \frameworkname's backend has identified. This feature is based on our algorithms for detection, re-identification, and tracking. \frameworkname maintains a database of the Eliteserien 2023 league's teams and players. Selecting a team from the API call presents a list of players matching our OCR detection. Users can then choose their preferred player for video clipping. Additionally, by selecting a team, users can view the home and away jerseys on the left sidebar, and by selecting a player, the player's photo is displayed. This visualize the player which we are going to track and clip videos.

\end{itemize}

After selecting the desired options, pressing the "Clip Video" runs the \frameworkname core pipeline, and ultimately clips the input video based on the selected player, showcasing only the relevant frames. 

\begin{table}[h!]

    \centering
    \footnotesize
    
    \begin{tabular}{|c|l|l|}
    
        \hline
        \textbf{Module} 
        & \textbf{Configurable Parameters} 
        & \textbf{Options}
        \\ \hline


        \multirow{4}{*}{Tracking} 
        & Object detection model
        & YOLOX, YOLOv8 
        \\ \hhline{~--} 
        
        & DeepEIoU parameters
        & (mult.) 
        \\ \hhline{~--} 

        & Appearance threshold
        & Default: $0.25$
        \\ \hhline{~--} 

        & Obj. det. confidence threshold
        & Default: $0.25$
        \\ \hline 


        \multirow{6}{*}{RGB} 
        & Color space
        & RGB, CIELAB, Weighted RGB 
        \\ \hhline{~--} 
        
        & Clustering method
        & KMeans
        \\ \hhline{~--}

        & Offset type
        & trim, center, none
        \\ \hhline{~--}
        
        & Normalization
        & TRUE/FALSE 
        \\ \hhline{~--}

        & Scaling
        & TRUE/FALSE 
        \\ \hhline{~--}

        & \gls{iou} threshold
        & [0, 1] 
        \\ \hline 

        \multirow{4}{*}{OCR} 
        & \gls{ocr} method
        & PaddleOCR, EasyOCR 
        \\ \hhline{~--} 

        & Crop score function
        & iou\_score, iou\_area, iou\_random 
        \\ \hhline{~--} 

        & Get \gls{ocr} function
        & top \textit{n}, split into time zones
        \\ \hhline{~--}
        
        & Stop at detection
        & TRUE/FALSE
        \\ \hline

        
    \end{tabular}

    \caption{\frameworkname core pipeline selected parameters.}
    \label{tab:config-parameters}
    
\end{table}


\subsubsection{Output}

\begin{figure}[htbp]
    \centering
    \frame{\includegraphics[width=\columnwidth]{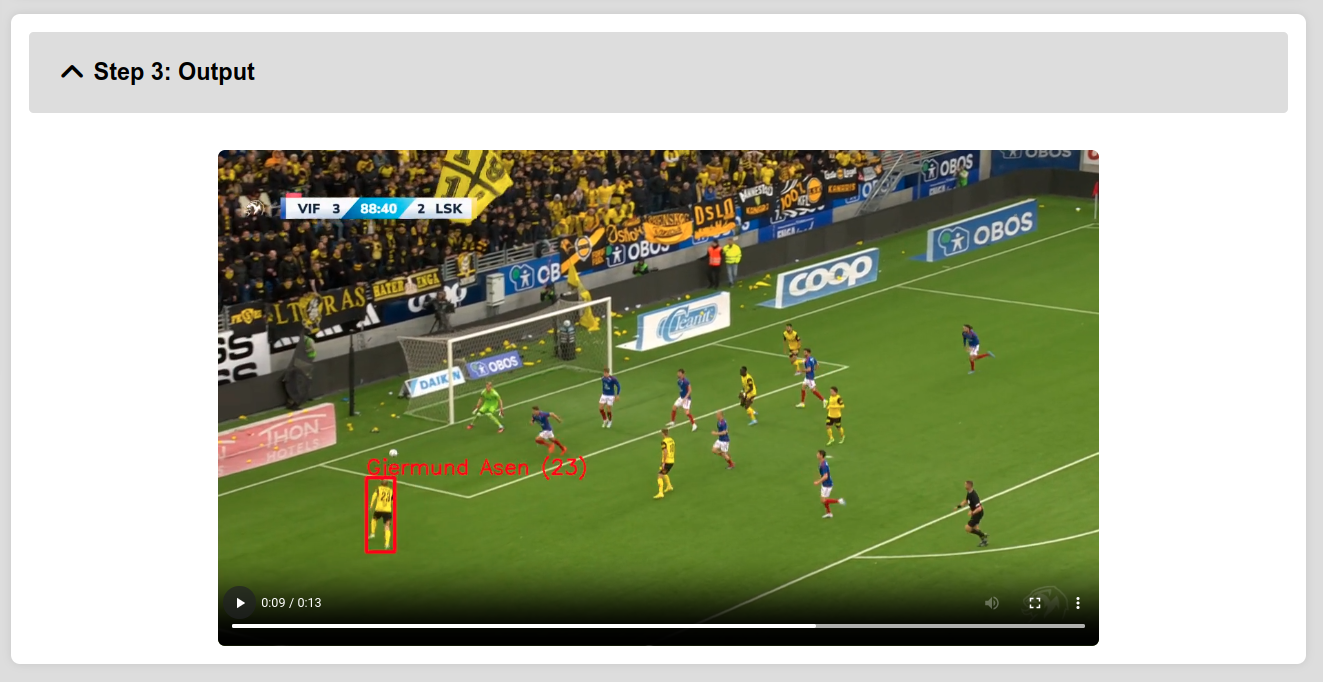}}
    \caption{\frameworkname \gls{gui} step 3.}
    \label{fig:demo-3}
\end{figure}

The output of this step is an MP4 file, which retains the same frame rate (FPS) as the original video. This file exclusively displays the frames where the selected player is detected and tracked. In these frames, the player is highlighted with a bounding box. Additionally, the player's name and jersey number are displayed above this bounding box for easy identification (see Figure~\ref{fig:demo-3}).

\subsection{Open Artifacts}

The entire codebase for the \frameworkname framework, including the core pipeline and the interactive \gls{gui} is publicly available as open source software under~\cite{PlayerTV-GitHub}. 
The repository also includes additional assets (team and player lookup tables), as well as a README which provides an overview of the software and hardware requirements, technical implementation details, and instructions on how to replicate the complete framework in an end-to-end manner. A video demonstration of the interactive \gls{gui} is available on YouTube~\cite{PlayerTV-YouTube}.


\section{Evaluation}\label{section:evaluation}

\begin{table*}[ht!]
    \centering

    \begin{tabular}{|c|c|c|c|c|c|c|}
    
        \hhline{~------}
        \multicolumn{1}{c}{}
        & \multicolumn{2}{|c|}{\textbf{Team Mapping Performance (\%)}} 
        & \multicolumn{2}{c|}{\textbf{OCR Performance (\%)}} 
        & \multicolumn{2}{c|}{\textbf{Combined (\%)}} 
        \\ \hhline{~------} 
        
        \multicolumn{1}{c|}{}
        & \textbf{W. RGB} 
        & \textbf{CIELAB} 
        & \textbf{Paddle} 
        & \textbf{Easy} 
        & \textbf{Paddle}
        & \textbf{Easy}
        \\ \hline 
                
        Video 1
        	& 55.6 / 66.7
        	& 77.8 / 85.2
        	& 23.8 (33.3)
        	& 14.3 (14.3)
        	& 23.8
        	& 14.3 
         \\ \hline
        
        Video 2
        	& 56.8 / 70.3
        	& 83.8 / 86.5
        	& 39.1 (43.5)
        	& 17.4 (17.4)
        	& 39.1
        	& 17.4 \\ \hline
        
        Video 3
        	& 96.7 / 100.0
        	& 96.7 / 100.0
        	& 20.0 (20.0)
        	& 0 (0)
        	& 20.0
        	& 0.0 \\ \hline
        
        Video 4
        	& 95.0 / 100.0
        	& 95.0 / 100.0
        	& 41.2 (50.0)
        	& 5.9 (5.9)
        	& 41.2
        	& 5.9 \\ \hline
        
        Video 5
        	& 94.9 / 100.0
        	& 92.3 / 100.0
        	& 40.0 (40.0)
        	& 10.0 (10.0)
        	& 40.0
        	& 10.0 \\ \hline
        
        Video 6
        	& 93.8 / 93.8
        	& 100.0 / 100.0
        	& 46.2 (53.8)
        	& 0.0 (7.7)
        	& 46.2
        	& 0.0 \\ \hline
        
        Video 7
        	& 98.0 / 100.0
        	& 98.0 / 98.0
        	& 29.2 (29.2)
        	& 4.2 (4.2)
        	& 29.2
        	& 4.2 \\ \hline
        
        Video 8
        	& 94.6 / 94.6
        	& 94.6 / 94.6
        	& 22.7 (22.7)
        	& 4.5 (4.5)
        	& 22.7
        	& 4.5 \\ \hline
        
        \textbf{Video 9}
        	& \textbf{76.2 / 71.4}
        	& \textbf{100.0 / 95.2}
        	& \textbf{0 (0)}
        	& \textbf{0 (0)}
        	& \textbf{0}
        	& \textbf{0} \\ \hline
        
        Video 10
        	& 97.7 / 100.0
        	& 90.7 / 95.3
        	& 12.5 (12.5)
        	& 6.2 (6.2)
        	& 12.5
        	& 6.2 \\ \hline
        
        Video 11
        	& 100.0 / 100.0
        	& 100.0 / 100.0
        	& 38.5 (38.5)
        	& 7.7 (7.7)
        	& 38.5
        	& 7.7 \\ \hline
        
        Video 12
        	& 100.0 / 100.0
        	& 100.0 / 100.0
        	& 42.9 (50.0)
        	& 17.9 (17.9)
        	& 42.9
        	& 17.9 \\ \hline
        
        Video 13
        	& 100.0 / 96.2
        	& 96.2 / 96.2
        	& 27.8 (27.8)
        	& 5.6 (11.1)
        	& 27.8
        	& 5.6 \\ \hline
        
        Video 14
        	& 92.9 / 100.0
        	& 100.0 / 100.0
        	& 62.5 (75.0)
        	& 62.5 (62.5)
        	& 62.5
        	& 62.5 \\ \hline
        
        Video 15
        	& 85.0 / 90.0
        	& 100.0 / 90.0
        	& 33.3 (33.3)
        	& 11.1 (22.2)
        	& 33.3
        	& 11.1 \\ \hline
        
        Video 16
        	& 87.5 / 87.5
        	& 100.0 / 93.8
        	& 33.3 (33.3)
        	& 33.3 (33.3)
        	& 33.3
        	& 33.3 \\ \hline
        
        \textbf{Overall} 
        & 87.6 / 89.8
        & 91.5 / 93.7 
        & 30.6 (35.7)
        & 11.3 (12.3)
        & 29.8
        & 11
        \\ \hline
    \end{tabular}
    \caption{\frameworkname evaluation.}
    \label{tab:evaluation-combo}
\end{table*}

\begin{figure}[t]
    \centering
      \includegraphics[width=\columnwidth]{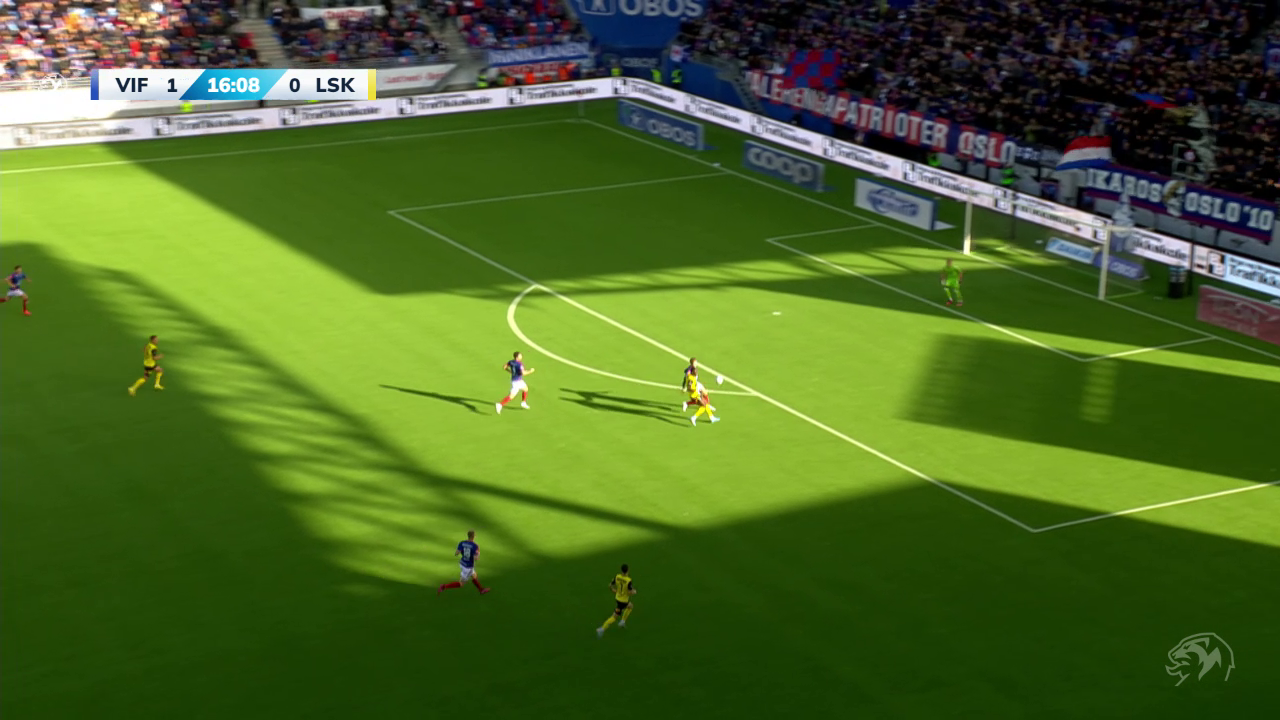} 
    \caption{Sample frame from Video 9 - Challenging light conditions make weighted RGB perform worse than CIELAB.}\label{fig:rgb-vs-cielab}
\end{figure}

For evaluating the \frameworkname core pipeline, we have taken the tracker output as baseline for tracklet ground truths. Running through all tracklets in a 900-frame sequence video, before manually labeling each with team name as well as kit number. To ensure the integrity of the data annotated, filtering was applied. We retained only those player tracklets that consistently maintained focus on a single player across frames. Additionally, not all tracklets have a visible kit number throughout the clip. Such tracklets have been labeled "-1" and are excluded from the OCR performance metrics. A total of 16 videos from the Norwegian top division "Eliteserien" have been manually annotated. This resulted in a sample base consisting of 511 instances of player tracklets, each mapped to a specific team. A subset of these 511 tracklets have visible kit numbers throughout the clip, resulting in 300 tracklets annotated with both team and kit number. 

Dataset creation and evaluation process is as follows: 
\begin{enumerate}
    \item Default Deep-EIoU tracker is run for clips, providing a file with tracklet information.
    \item Each individual tracklet with tracking id is mapped to correct team and number. Inconsistent tracklets are discarded. This results in sequences of crops of varying lengths, each consistently featuring the same player. 
    \item By combining these mappings, we evaluate the RGB module and OCR module individually, as well as their combination.\\ \textbf{RGB Module:} Accuracy measurement for correctly mapped team from player tracklet. Both for CIELAB and weighted RGB color space.\\ \textbf{OCR Module:} Accuracy as well as top-5 accuracy (in parentheses) for each player tracklet.\\ \textbf{Combined:} Accuracy where both team and kit number are correct for a tracklet. 
    \item We evaluate the same metrics for post processed results as well as results achieved in online mode. Table~\ref{tab:evaluation-combo} presents the results of our evaluation, combined columns indicate PaddleOCR and EasyOCR with CIELAB.
\end{enumerate}


\subsection{RGB Module Performance}

This module is the most accurate. Not surprising considering its task. Using the CIELAB color space also improved correct team labeling by 3.9\% compared to using the weighted RGB metric. Post processing the results, increased performance by an additional 2\%. These findings suggests that relatively straightforward measures, as mean color value, could be effective. This approach could serve as a robust alternative to employing more complex machine learning modules for processing player crops. Figure~\ref{fig:rgb-vs-cielab} shows an example frame from Video 9 in Table~\ref{tab:evaluation-combo}, revealing the limitations of the weighted RGB approach. In the challenging light conditions, the CIELAB approach is perfect for the online clustering, while the weighted RGB struggles having an accuracy of only 76.2.


\subsection{OCR Module Performance}

The main constraint within this pipeline is the kit association, which peaks at just over 30\%. The the gap in performance between the EasyOCR vs PaddleOCR models is notable, with PaddleOCR outperforming by a factor of three. Specifically, within our dataset of 300 tracklets, the PaddleOCR model, ranks the correct kit number within the top 5 detections for 107 instances. In 92 of these 107 instances, the correct number appears as the top-ranked detection. To further refine our results, reducing the disparity between these measures is crucial. A possible improvement could involve assigning additional weight to the detection's position within the crop itself.

Given that both the EasyOCR and PaddleOCR models were initially developed for purposes slightly different from their current usage in this pipeline, training a custom detector in the future is desirable. Another factor to consider when reading the OCR performance is also the length of the clips provided. If clips span over a longer period of time, the OCR module will have more chances of detecting a kit number for a tracklet, improving performance.


\subsection{Team and Player Mapping Performance}

As the OCR detection module is the limiting factor, we see that the combined performance is marginally worse than the OCR module independently. Given this situation, it is reasonable to assume that enhancements in OCR accuracy would lead to proportional improvements in the pipeline's overall performance. Improvements for the OCR component could therefore significantly improve overall performance.


\subsection{System Performance}

In order to evaluate system performance, we consider the end-to-end pipeline processing speed. The \frameworkname pipeline was run on a computational cluster known as ex3\footnote{\url{https://www.ex3.simula.no/}}. Performance measures was carried out on a node equipped with an NVIDIA Tesla V100-SXM3 GPU. It is worth noting that the traffic on this cluster varies, influencing performance slightly. Our pipeline, using a greedy OCR approach that involves running EasyOCR across all crops in all frames, produced results at a rate of approximately 1 frame per second (fps).


\section{Conclusion}\label{section:conclusion}

We have introduced \frameworkname, a framework which combines technologies such as object detection and tracking, text recognition, and color analysis to automate the creation of player-specific video highlights. Our evaluations show promising accuracy in player and team identification, which substantiates the framework's utility in professional sports analytics. As we continue to refine the technology and expand its applications, \frameworkname is poised to transform how sports footage is processed and utilized, enhancing both the strategic elements of sports team management and the overall fan experience.

\section*{Acknowledgment}

This research was partly funded by the Research Council of Norway, project number 346671 (AI-Storyteller), and has benefited from the Experimental Infrastructure for Exploration of Exascale Computing (eX3), which is financially supported by the Research Council of Norway under contract 270053. The authors would like to thank the Norwegian Professional Football League ("Norsk Toppfotball") for making videos available for research. 

\balance
\bibliographystyle{IEEEtran}
\bibliography{references}

\end{document}